\documentclass[journal]{IEEEtran}
\usepackage{amsmath,amsfonts}
\usepackage{algorithmic}
\usepackage{algorithm}
\usepackage{array}
\usepackage[caption=false,font=normalsize,labelfont=sf,textfont=sf]{subfig}
\usepackage{textcomp}
\usepackage{stfloats}
\usepackage{url}
\usepackage{verbatim}
\usepackage{graphicx}
\usepackage{cite}
\usepackage{url}
\usepackage{hyperref}
\usepackage{color}
\usepackage{xcolor}
\usepackage{xcolor,graphicx}
\usepackage[center]{caption}
\usepackage[none]{hyphenat}
\ifCLASSINFOpdf
\else
\fi
\usepackage{algorithmic}

\usepackage{array}

\begin{document}
%
\title{\textbf{Gen AI in Automotive: Applications, Challenges, and Opportunities with a Case study on In-Vehicle Experience}}

%
%
%

\author{Chaitanya Shinde, Divya Garikapati}

\maketitle
\begin{abstract}
\textbf{
Generative Artificial Intelligence (GenAI) is emerging as a transformative force in the automotive industry, enabling novel applications across vehicle design, manufacturing, autonomous driving, predictive maintenance, and in-vehicle user experience. This paper provides a comprehensive review of the current state of GenAI in automotive, highlighting enabling technologies such as Generative Adversarial Networks (GANs) and Variational Autoencoders (VAEs). Key opportunities include accelerating autonomous driving validation through synthetic data generation, optimizing component design, and enhancing human–machine interaction via personalized and adaptive interfaces. At the same time, the paper identifies significant technical, ethical, and safety challenges, including computational demands, bias, intellectual property concerns, and adversarial robustness, that must be addressed for responsible deployment. A case study on Mercedes-Benz’s MBUX Virtual Assistant illustrates how GenAI-powered voice systems deliver more natural, proactive, and personalized in-car interactions compared to legacy rule-based assistants. Through this review and case study, the paper outlines both the promise and limitations of GenAI integration in the automotive sector and presents directions for future research and development aimed at achieving safer, more efficient, and user-centric mobility. Unlike prior reviews that focus solely on perception or manufacturing, this paper emphasizes generative AI in voice-based HMI, bridging safety and user experience perspectives.}

\end{abstract}


\section{Introduction}\label{introduction}
%
%
%
%

The integration of Artificial Intelligence (AI) into the automotive industry has been a gradual but transformative journey \cite{akande2024review,ben2023context,bendoly2023role,doi:10.1049/PBTR046E_ch4}. In the early years, AI had very limited applications within the manufacturing plants \cite{casteleiro2024generative} or static applications where human error could be improved like fixed robots performing mundane tasks but with precision. With the introduction of Advanced Driver Assistance Systems (ADAS) \cite{choe2023emotion}, which included features such as adaptive cruise control, lane keeping assist etc., the role of AI and related technologies evolved to more dynamic applications like Automated Driving. The ADAS systems paved the way for more direct interaction of AI with vehicles making them more intelligent to assist the driver as needed. Then with the vast improvements in the sensing technologies, AI-powered vehicles became even more a reality. Companies like Tesla, Google (Waymo), and Uber lead the way in implementing these technologies to make them more tangible for real operation of intelligent vehicles in public road environments. This progression demonstrates how each technological wave, manufacturing robotics, driver assistance, and now generative intelligence, has shifted the locus of AI innovation closer to direct driver interaction and experience.

Currently, AI is not only used for driving automation but also in areas like predictive maintenance, prognostics, in-car virtual assistants, and enhanced user experiences within the vehicles. More and more features are being added that use AI and data-driven approaches to perform the driving and non-driving related vehicle tasks. Millions of miles of data is being collected by the automated vehicles to be able to use the data to train the AI systems.

There are multiple types of AI technologies that vary in complexity and capability. At the basic level, reactive machines respond to specific stimuli, while limited memory AI can use past experiences to inform decisions. More advanced AI, like theory of mind and self-aware AI, remain largely theoretical, focusing on understanding human emotions and self-awareness. AI is often categorized into Artificial Narrow Intelligence (ANI), which is task-specific, Artificial General Intelligence (AGI), which aims to replicate human cognitive abilities, and Artificial Superintelligence (ASI), a speculative concept surpassing human intelligence. Key AI technologies include machine learning (ML) and its subset deep learning (DL), which drive many modern applications, as well as generative AI, which creates new content. Other AI forms include expert systems, which emulate human expert decision-making, and natural language processing (NLP), which enables machines to understand and process human language.

Generative AI refers to a class of artificial intelligence models designed to generate new content, whether images, text, audio, or even designs, that are indistinguishable from real data. Unlike traditional AI models that are typically used for classification, detection, or prediction tasks, generative AI creates novel outputs based on the patterns it has learned from existing data. In the automotive industry, generative AI has far-reaching implications. For instance, it can be used to create realistic simulations of driving environments for testing autonomous vehicles, generate synthetic data to train AI models, and even design new vehicle components or optimize manufacturing processes. By leveraging generative AI, the automotive industry can accelerate the development of safer, more efficient, and innovative vehicles, reducing the time and cost associated with traditional design and testing methods.

Generative AI is powered by several key technologies, among which Generative Adversarial Networks (GANs) and Variational Autoencoders (VAEs) are the most prominent.

Generative Adversarial Networks (GANs): Introduced by Ian Goodfellow in 2014, GANs consist of two neural networks—the generator and the discriminator, that are trained simultaneously. The generator creates fake data, while the discriminator evaluates its authenticity. Through this adversarial process, the generator improves its ability to produce realistic data over time. GANs have been used in the automotive industry for generating synthetic driving data, enhancing image quality, and simulating different driving scenarios.

Variational Autoencoders (VAEs): VAEs are a type of autoencoder that introduces a probabilistic element to the encoding process. Unlike traditional autoencoders, which map input data to a latent space and then reconstruct it, VAEs assume the latent space follows a certain distribution (usually Gaussian). This allows VAEs to generate new data points by sampling from this latent space. In the automotive sector, VAEs are used for anomaly detection, predictive maintenance, and generating variations of vehicle designs.

These technologies form the backbone of generative AI in the automotive industry, enabling the creation of high-fidelity simulations, enhanced design processes, and more efficient development workflows. The preceding discussion outlines the technological foundation necessary for integrating generative AI into vehicle development. The following section consolidates the key contributions and scope of this work.

\subsection{Contributions of this Paper:}

This paper aims to explore the impact of generative AI on the automotive industry, focusing on its applications, benefits, and potential challenges. The objectives are to provide a comprehensive understanding of how generative AI technologies like GANs and VAEs can be leveraged to innovate and enhance various aspects of automotive design, manufacturing, and autonomous driving. The paper is structured as follows:

\begin{enumerate}
    \item \textbf{Introduction}: A brief history of AI in automotive applications and an introduction to generative AI.
    \item \textbf{Generative AI Technologies}: An in-depth look at key generative AI technologies, including GANs and VAEs, and their relevance to the automotive industry.
    \item \textbf{Applications in the Automotive Industry}: Exploration of specific use cases, such as Voice Assistant in Automotive Infotainment 
    \item \textbf{Challenges and Future Directions}: Discussion of the challenges in implementing generative AI and potential future developments in this space.
    \item \textbf{Conclusion}: A summary of key findings and the potential impact of generative AI on the future of the automotive industry.
\end{enumerate}

\section{Challenges}
\subsection{Technical Challenges}
The integration of Generative AI within the automotive industry presents a host of technical challenges that must be addressed to ensure effective deployment. One significant issue is the high computational costs associated with training and deploying Generative Adversarial Networks (GANs), particularly in the domain of autonomous vehicles where real-time decision-making is critical \cite{chaudhari2024generative}. Additionally, the application of Generative AI in radar systems highlights challenges related to data accuracy and the need for advanced algorithms that can handle the dynamic and unpredictable nature of automotive environments \cite{cakan2024}. The need for accurate and high-quality data to train generative models is another major hurdle. In the automotive domain, data quality issues can lead to unreliable model predictions and system behaviors, thus requiring robust data preprocessing and validation strategies to ensure the reliability of AI-driven systems \cite{li2024ai}. Moreover, latency issues and the difficulty of performing distributed inference across heterogeneous vehicle networks pose additional barriers to the seamless integration of these technologies  \cite{10634792}. Lastly, while Generative AI has shown potential in enhancing vehicle design and diagnostics, its effectiveness is often limited by infrastructure constraints and the accuracy of the models, necessitating further research to overcome these limitations \cite{dash2024ai}

\subsection{Ethical and Legal Challenges}

The potential for bias in AI-generated outcomes is also a significant concern, as biased data can lead to unfair or discriminatory decisions in critical areas such as autonomous driving and safety assessments \cite{10634792}. Furthermore, intellectual property concerns arise with AI-generated designs, particularly regarding the ownership and protection of AI-created innovations, which could lead to legal disputes and challenges in the automotive industry \cite{thuraisingham2024trustworthy}.

\subsection{Safety and Security Challenges}

The application of Generative AI in the automotive industry brings forth significant safety and security challenges that must be meticulously addressed \cite{andreoni2024enhancing,krstavcic2024safety}. One of the primary concerns is the potential for adversarial attacks that exploit vulnerabilities in AI models, which can lead to unsafe decisions in autonomous vehicles, thereby compromising passenger safety \cite{susmitha2023intricate}. Additionally, the ethical and security implications of deploying AI systems in vehicles are profound, as failures or security breaches in these systems could result in catastrophic outcomes. The integration of Generative AI also raises concerns regarding the robustness of attack detection systems in smart cars, where ensuring the trustworthiness of AI-driven decisions is critical to maintaining security and safety standards \cite{thuraisingham2024trustworthy}. Moreover, the deployment of Generative AI in autonomous systems necessitates rigorous safety validation, as inadequately validated AI models may fail to recognize or appropriately respond to complex real-world scenarios, thereby posing serious safety risks \cite{decker2023towards}. Kirchner and Knoll (2025) demonstrate that LLM-based automotive code generation can be integrated with formal verification loops for functions such as Adaptive Cruise Control, illustrating both the potential and risk of generative AI in safety-critical software\cite{kirchner2025generating}. Furthermore, the increasing reliance on AI for vehicle safety and security highlights the need for resilient and secure AI architectures that can withstand emerging threats and ensure the continuous safety of passengers and pedestrians alike \cite{andreoni2024enhancing}. Lastly, the ethical and security implications of deploying Generative AI in vehicles are profound, particularly as these systems become more autonomous. A failure or breach in these AI-driven systems could result in catastrophic outcomes, raising serious concerns about the trustworthiness of AI in ensuring the safety of passengers and pedestrians \cite{garikapati2024autonomous}. Addressing these challenges is crucial to the safe and secure integration of Generative AI in the automotive industry.

\section{Opportunities}

The application of Generative AI in the automotive industry offers numerous promising opportunities that can revolutionize various aspects of vehicle design, manufacturing, and operation. One of the most significant opportunities lies in enhancing vehicular networks, where Generative AI can optimize data transmission, improve network efficiency, and enable more reliable communication between autonomous vehicles \cite{teef2024enhancing}. Additionally, Generative AI provides innovative solutions for the development of AI-driven auto-scaling systems, which can monitor and adapt to changing conditions in real-time, thereby improving vehicle performance and safety \cite{yadav13optimizing}. The technology also plays a crucial role in advancing autonomous driving simulations, where AI-driven models can generate realistic scenarios and pedestrian behaviors, significantly enhancing the training and validation processes for autonomous systems \cite{ramesh2024walk}. Complementing these efforts, Da et al. (2025) provide a comprehensive survey of generative AI methods in transportation planning, mapping retrieval-augmented generation, diffusion, and zero-shot learning techniques to urban-mobility applications \cite{da2025generative}.

Moreover, the ability of Generative AI to create synthetic data is another key opportunity, as it enables the automotive industry to overcome the challenges of limited real-world data availability. This capability allows for the generation of diverse datasets that can be used to train AI models more effectively, leading to improved accuracy and robustness in vehicle systems \cite{thielen2024comparative}. In the context of vehicle electrification, Generative AI can also assist in optimizing battery management systems and enhancing the efficiency of energy storage solutions, contributing to the broader goals of sustainable transportation \cite{li2024ai}. Additionally, the integration of Generative AI in software development processes, particularly in areas such as prompt engineering, presents opportunities for streamlining regulatory compliance and accelerating innovation within the automotive sector \cite{backlund2024exploring}.

These opportunities demonstrate the transformative potential of Generative AI in driving innovation and enhancing the capabilities of the automotive industry, paving the way for more efficient, safer, and sustainable transportation solutions. Collectively, these opportunities signal the onset of an innovation cycle in which generative AI supports not only performance optimization but also safety assurance and user trust across the vehicle lifecycle.

\section{Gen AI Applications Literature Review} \label{literature_review}

To establish a comprehensive understanding of the current research landscape, this section reviews recent developments in generative AI applications, beginning with general cross-domain use cases and then focusing specifically on the automotive sector. The objective is to identify how foundational advances in generative modeling—such as those in image synthesis, text generation, and simulation—are being adapted to address domain-specific challenges in vehicle design, autonomous driving, predictive maintenance, and human–machine interaction. By examining both foundational and applied studies, the literature review highlights emerging trends, methodological innovations, and critical research gaps that define the trajectory of generative AI integration within the automotive industry. For example, Pan et al. (2025) propose a framework that combines LLM-based requirement interpretation with formal model generation to automate parts of the automotive software lifecycle in safety-critical systems \cite{pan2025automating}. Among the reviewed studies, four primary research clusters emerge, i.e., (i) design and manufacturing, (ii) autonomous driving simulation, (iii) predictive maintenance and monitoring, and (iv) user experience design.

\subsection{Applications of Generative AI in the Automotive Industry}

\subsubsection{Design and Manufacturing}

    The study by Ishihara et al. (2023) \cite{ishiharakansei} explores the integration of generative AI into the Kansei engineering design process, particularly for automotive headlight design and demonstrate how AI enables precise design customization by allowing designers to specify and manipulate specific areas, significantly enhancing the representation of proposed models. Li et al. (2024) \cite{li2024advancing} examine how generative AI is transforming automotive design processes, particularly by enhancing production efficiency and creativity to highlight how AI-driven tools streamline the design workflow, leading to more innovative and efficient design outcomes. According to Gazi (2024) \cite{gazi2024implications}, it is predicted that AI generative and machine learning will be at the forefront of the evolution in the automotive industry, particularly toward the reduction of its carbon print. Bendoly et al. (2023) \cite{bendoly2023role} discuses the role of generative design in additive manufacturing and a potential symbiosis between human and AI collaborative design. Akande et al. (2024) \cite{akande2024review} critically reviewed the use of generative models, including GANs and VAEs, to automate and optimize the design and simulation of 3D vehicle wheels. The attention here is on bringing to light the transformational potential of infusing deep learning into the traditional CAD/CAE system for improved design processes and accurate simulations. Elrefaie et al. (2025) introduce a multi-agent generative-design framework that links language, vision, and geometric reasoning models to accelerate car body styling and aerodynamic optimization, reducing iterative design cycles from weeks to minutes\cite{elrefaie2025ai}. Doanh et al. (2023) explain the role of GAI in predictive maintenance processes for various manufacturing activities. It is important to note that GAI predicts maintenance requirements through analyzing data and predicting equipment breakdown before the scheduled period, thus enhancing operational efficiency and the minimization of downtime.

\subsubsection{Autonomous Driving}

The recent approach of applying generative AI to vehicular mixed reality metaverses by Xu et al. (2024) \cite{xu2023generative}presents the ability to simulate difficult driving scenarios. The work demonstrates how such simulations could help polish and test the autonomous driving systems under varied conditions. Gaia-1, in contrast, was offered for autonomous driving prediction and simulation of scenarios by Hu et al. (2023) \cite{hu2023gaia}through a generative AI-driven world model; emphasis was laid on the importance of realistic environment simulation for training and validation. Marathe et al. (2023) \cite{marathe2023wedge}propose Wedge, a novel dataset synthesized from artificial intelligence models that focus on multiweather conditions for autonomous driving. It brings forth important insights towards how one can raise the bar even further in terms of robustness and accuracy for autonomous systems under different weather condition

\subsubsection{Maintenance and Monitoring}

Liu et al. (2021) \cite{liu2021novel}present a method leveraging \textbf{LSTM-GAN} for self-detection of abnormal data and predictive maintenance in intelligent manufacturing systems. Their approach improves machine reliability by predicting faults before they occur, using GAN to model diverse machine degradation scenarios.- Predictive maintenance through anomaly detection. Ucar et al. (2024)\cite{ucar2024artificial} review recent developments in AI-based predictive maintenance, focusing on the integration of AI with data analytics to enhance system autonomy and predictive accuracy. The authors emphasize the role of generative models in simulating rare or unseen failure scenarios for enhancing maintenance predictions. In their findings, Prabhod et al. (2021) \cite{prabhod2021advanced} explored the applications of G-AI and the use of deep learning models combined with CNNs and RNNs for real-time monitoring. He used G-AI to generate realistic sensor data that augments datasets in improving failure prediction models in industrial IoT scenarios. Doanh et al. (2023) \cite{doanh2023generative}explain the role of GAI in predictive maintenance processes for various manufacturing activities. It is important to note that GAI predicts maintenance requirements through analyzing data and predicting equipment breakdown before the scheduled period, thus enhancing operational efficiency and the minimization of downtime.

\subsubsection{In-Vehicle Experience and UI/UX Design}

Choe et al. (2023) \cite{choe2023emotion} study the integration of generative AI technologies such as ChatGPT, DALL·E 2 in developing empathic in-vehicle interfaces. The tools are used for creating contextual emotional responses within automated driving scenarios, supporting user-vehicle interaction enhancement by enabling real-time emotional regulation and facilitating fostered user trust toward the vehicle. Casteleiro-Pitrez (2024) \cite{casteleiro2024generative}has evaluated the usability and emotional impact of generative AI image tools, such as Midjourney, DreamStudio, and Adobe Firefly. From that study, he showed that the tools produce positive results in efficiency and ease of learning but fall short in novelty and stimulation, which indicates improvement in engaging user experiences. Uusitalo et al. (2024) \cite{uusitalo2024clay} explore how generative AI tools, such as Midjourney and ChatGPT, are used in UX and industrial design for storytelling, ideation, and visual consistency. Designers use these tools to overcome creative blocks, enhance collaboration, and generate moodboards and personas, though concerns over control and ownership persist. Lazzaroni et al. (2024) \cite{lazzaroni2024facets} describes an end-to-end voice assistant system, which is optimized for vehicles and designed to work off-line. In this study, privacy and security features are focused on bypassing the cloud-based solutions by using edge computing. The assistant is based on natural language understanding, automatic speech recognition, and speech synthesis to increase real-time interaction in the automotive context.

\section{In-Vehicle Experience – Gen AI application in Voice assistance}

Among all automotive use cases, in-vehicle voice assistance provides the most direct interface between generative AI and human perception of vehicle intelligence. Voice assistance represents one of the most visible and user-centric applications of artificial intelligence, enabling natural language interaction between humans and machines. Broadly, voice assistants can be categorized into two domains: in-home and in-vehicle. In-home assistants, such as Amazon Alexa, Google Assistant, and Apple Siri, operate in relatively stable, low-noise environments with persistent network connectivity, focusing on information retrieval, entertainment, and smart-home control. In contrast, in-vehicle voice assistants must function under far more dynamic conditions characterized by background noise, motion, variable connectivity, and stringent safety requirements. This context demands specialized architectures that combine automatic speech recognition (ASR), natural language understanding (NLU), dialogue management, and increasingly, generative AI models capable of contextual reasoning and personalization. The emergence of large language models (LLMs) and multimodal generative systems has elevated in-vehicle assistants from basic command execution to adaptive copilots that anticipate driver needs, provide proactive support, and integrate seamlessly with vehicle functions. This section explores how generative AI reshapes the in-vehicle voice-assistance landscape, emphasizing the transition from rule-based interaction to intelligent, conversational collaboration.

Some of the features of modern day voice assistants: 

\begin{enumerate}
    \item \textbf{Natural conversations}: Gen AI enables more human-like interactions with in-car voice assistants. For example, Mercedes-Benz's upgraded MBUX Virtual Assistant uses generative AI to make conversations more natural and personalized, offering reliable and relevant responses.
    \item \textbf{Personalization: }Drivers can customize their voice assistants to suit their preferences. Tesla, for instance, allows owners to customize the tone and speaking style of their Grok intelligent assistant.
    \item \textbf{Expanded functionality: }Gen AI-powered assistants can handle complex tasks beyond simple commands. They can access vital vehicle information, control various features, and even participate in discussions about news, points of interest, or weather.
    \item \textbf{Proactive assistance: }These advanced assistants can anticipate user needs and offer proactive solutions based on context and past behavior.
    \item \textbf{Enhanced safety:} By enabling more natural voice interactions, Gen AI helps drivers stay focused on the road while accessing information or controlling vehicle functions. With additional vital sensors for drivers/ passengers.
    \item \textbf{Improved navigation}: AI assistants can provide real-time traffic updates, suggest alternative routes, and remind drivers about appointments while driving.
    \item \textbf{Predictive maintenance}: Gen AI can analyze vehicle data to predict potential issues and schedule maintenance, improving vehicle reliability and reducing downtime.
    \item \textbf{Customer support}: AI-powered chatbots can handle complex queries about vehicle specifications, pricing, and availability, streamlining the car buying process, and improving after-sales support.
    \item \textbf{Personalized recommendations}: Based on user interactions, Gen AI can offer tailored suggestions for vehicle features, upgrades, or services.
    \item \textbf{Integration with smart home systems}: Advanced voice assistants allow drivers to control their smart home devices from their vehicles, creating a seamless connected experience.
\end{enumerate}
The next section presents a case study illustrating how these capabilities materialize in production systems, comparing legacy and modern architectures.

\section{Case Study}
To demonstrate the practical impact of generative AI within the automotive domain, this section presents a focused case study on its application to in-vehicle voice assistance. While prior sections discussed the broader technological landscape, the following analysis illustrates how these advancements materialize in a production environment. Specifically, the case study examines Mercedes-Benz’s next-generation MBUX Virtual Assistant, an AI-powered system that integrates large language models and multimodal interaction within the Mercedes-Benz Operating System (MB.OS). The study explores how generative AI enables natural dialogue, personalization, and proactive support while enhancing safety and driver engagement. By contrasting this implementation with earlier, non-AI voice-assistant generations, the case study provides a tangible view of how generative AI transforms user interaction paradigms from scripted automation to intelligent, adaptive collaboration. This comparison framework uses both historical analysis and scenario-based evaluation to quantify the progression of capabilities. Each scenario was selected to reflect a common driver–vehicle interaction and to demonstrate measurable performance improvements enabled by generative AI.

\subsection{Non-AI voice assistants in automotive}

Older generation voice assistants in automotive settings, which did not utilize AI, primarily relied on basic voice recognition technology. These systems required users to speak predefined commands clearly and often struggled with background noise and accents. Here are some examples:

\begin{enumerate}
    \item Honda's Acura (2004): The first edition of Acura's voice control system required drivers to press a button on the steering wheel to activate listening mode. It allowed basic control over functions like temperature, calls, and navigation but had limitations in speech recognition and required a quiet environment to function effectively.
    \item Lexus GS Models (2006): Lexus introduced voice command features that were activated by pressing a button on the steering wheel. These features included navigation and basic audio and climate control commands. However, the system was short-lived due to poor performance and user dissatisfaction with speech recognition.
    \item Ford Sync (2007): Developed in partnership with Microsoft, Ford Sync allowed drivers to control Bluetooth-enabled phones and media players using voice commands. It required a quiet environment for accurate recognition and was limited to specific commands.
    \item Skoda Octavia (2013): Skoda's voice control system was connected via Bluetooth and had a restricted vocabulary and complex command structure, leading to user frustration. The system required reading prompts, which detracted from its safety benefits.
\end{enumerate}
These early systems were constrained by the technology of the time, lacking the AI-driven natural language processing capabilities that modern voice assistants possess. They often required precise command inputs and were less adaptable to variations in speech and environmental noise. \cite{kardome2024}

\begin{figure}
    \centering
     \includegraphics[width=\linewidth,keepaspectratio]{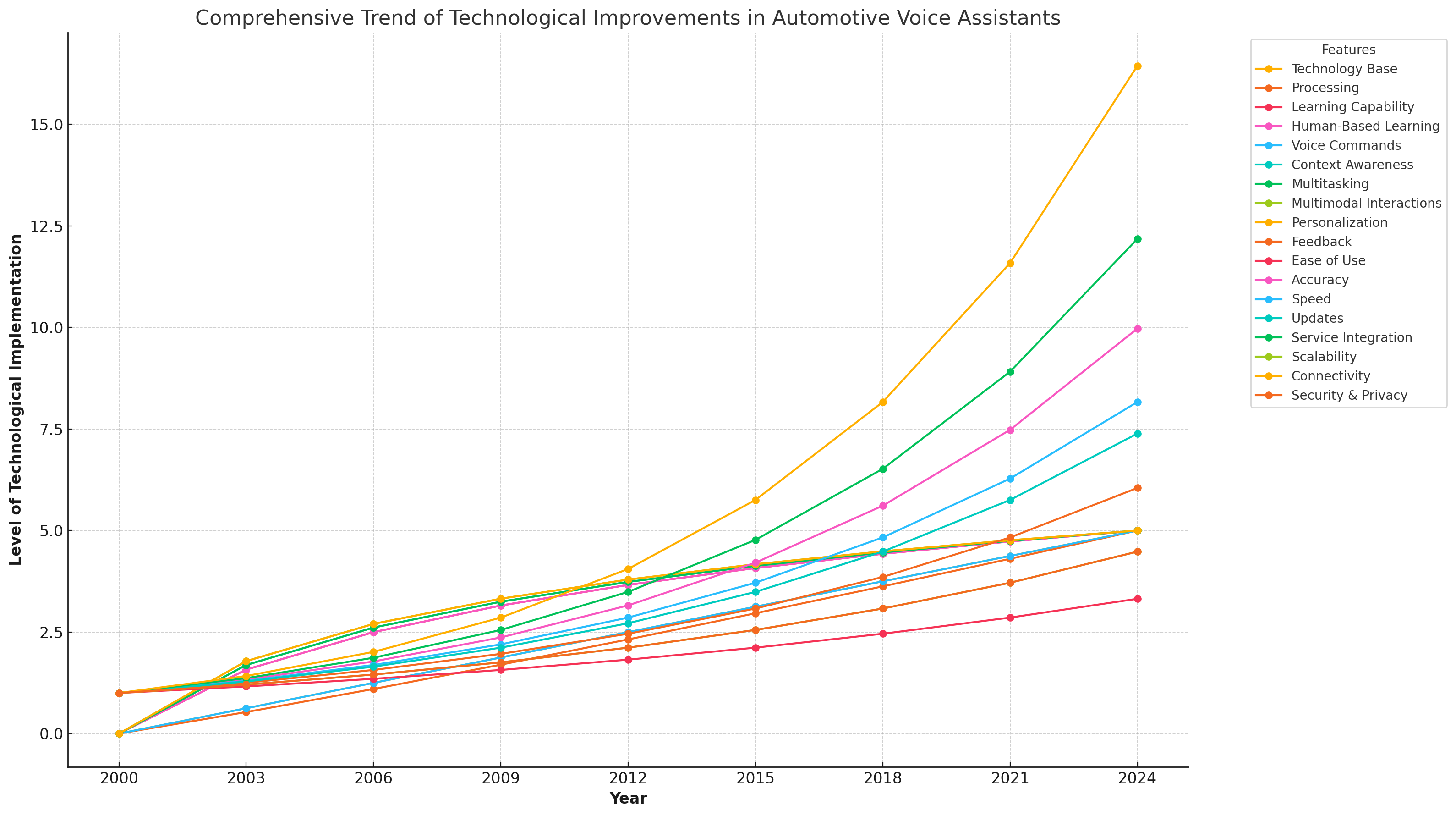}
    \caption{Voice Assistants trends over years in Automotive}
    \label{fig:Trends in automotive voice assistants}
\end{figure}

Fig.\ref{fig:Trends in automotive voice assistants} shows trends in Automotive Voice Assistants over time.
\subsubsection{\textbf{What were the main limitations of early car voice assistants?}}

\begin{enumerate}
    \item Limited Vocabulary and Command Structure: Early systems had a very restricted vocabulary, often limited to around 50-60 commands. Users needed to memorize specific commands or follow voice guidance to use the system, which limited flexibility and ease of use.
    \item Background Noise and Speech Recognition: These systems struggled with accurately recognizing speech in the noisy environment of a car. Background noise from the engine, wind, and other passengers often interfered with the system's ability to understand commands.
    \item Accent and Dialect Challenges: Variations in accents and speech patterns posed significant challenges for early voice recognition systems, leading to frequent misunderstandings and user frustration.
    \item Connectivity and Response Time: Early voice assistants often had slow response times and were dependent on poor or unreliable internet connections, which affected their functionality and user experience.
    \item Limited Integration and Features: These systems offered limited control over car functions and had difficulty integrating with third-party applications, restricting their usefulness.
    \item Privacy and Security Concerns: Even in early iterations, there were concerns about data privacy and security, as voice data could potentially be accessed or misused.\cite{dialzara2024, mobilityoutlook2024}
\end{enumerate}

Example: The Ford SYNC system is an integrated in-vehicle communications and entertainment system that allows drivers to make hands-free telephone calls, control music, and perform other functions with the use of voice commands. Here's a detailed review of the system focusing on its capabilities, user interface, and overall performance:

\subsubsection{\textbf{Overview}}

\begin{itemize}
    \item \textbf{Developer}: Ford Motor Company, in collaboration with Microsoft.
    \item \textbf{Launch}: Originally launched in 2007.
    \item \textbf{Purpose}: To enhance the driving experience by integrating mobile devices and allowing drivers to use voice commands for various controls.
\end{itemize}

\subsubsection{\textbf{Performance}}

\begin{itemize}
    \item \textbf{Reliability}: Early iterations had limitations in voice recognition accuracy, especially in noisy environments or with varied accents.
    \item \textbf{Speed}: Response times were generally adequate, though sometimes delayed, particularly with more complex commands or poor connectivity.
    \item \textbf{Updates}: Over the years, Ford released updates to improve functionality and expand features, including better voice recognition and more intuitive controls.
\end{itemize}

\subsubsection{\textbf{Pros}}

\begin{itemize}
    \item \textbf{Safety}: Reduced driver distraction by enabling hands-free operation of essential functions.
    \item \textbf{Convenience}: Allowed drivers to keep their hands on the wheel and eyes on the road while performing common tasks.
\end{itemize}

\subsubsection{\textbf{Cons}}

\begin{itemize}
    \item \textbf{Limited Command Flexibility}: Unlike AI-driven systems, SYNC required exact phrases and could not learn from user interactions or adapt to individual speech patterns.
    \item \textbf{Functionality Limitations}: Being non-AI, the system had a fixed set of capabilities and lacked the ability to process natural language or understand context.
\end{itemize}

\subsection{Actual Case Study Comparison}

The evolution of in-vehicle voice assistants provides a clear lens through which to examine the broader impact of generative AI on automotive user experience. Early systems such as Ford’s SYNC, launched in 2007, represented the first generation of voice-enabled interfaces that relied on fixed grammars and rule-based logic. While they improved driver convenience by enabling limited hands-free control, their functionality was constrained by vocabulary size, environmental noise sensitivity, and lack of contextual understanding. In contrast, modern assistants such as Mercedes-Benz’s MBUX leverage large language models and multimodal AI to interpret intent, maintain dialogue continuity, and personalize responses. To illustrate this technological progression, the following comparison evaluates three representative stages of development, Ford SYNC (rule-based), MBUX (LLM-powered production system), and advanced in-vehicle voice assistants currently under research. This case study highlights how generative AI transforms automotive human–machine interaction from predefined command execution to adaptive, context-aware collaboration.

\subsubsection{Comparison Ford SYNC vs MBUX vs Advanced In-Vehicle Voice Assistants (In Research)}

\begin{table}[htp]
\centering
\caption{Technology Foundation}
\fontsize{6.5pt}{7.5pt}\selectfont
\label{tab:standards}
\begin{tabular}{|p{1.2cm}|p{1.8cm}|p{2.3cm}|p{2.3cm}|}\hline


\textbf{Feature} & \textbf{Ford SYNC (2007)} & \textbf{MBUX (LLM-based)} & \textbf{Advanced In-Vehicle Voice Assistants (In Research)} \\
\hline
\textbf{Technology Base} & Rule-based command recognition & Large Language Models (AI-based natural language understanding) & Multi-modal AI \cite{wang2024multi}with contextual understanding and natural language processing\\
\hline
\textbf{Processing} & Local processing with limited external connectivity & Cloud-based processing with real-time updates and enhancements & Hybrid processing (edge and cloud) with real-time contextual awareness \cite{ben2023context} and low latency \cite{robert2023latency}\\
\hline
\textbf{Learning Capability} & None & Dynamic learning and adaptation based on user interaction & Continuous learning \cite{hurtado2023continual}from user interactions and environmental data, including predictive behavior adaptation \cite{maroto2023adaptive}. Autonomous Driving decisions communication based on behavior prediction \cite{fang2024behavioral} \cite{zhou2023research} \cite{song2023trust}. \\
\hline
\textbf{Human-Based Learning} & None & Limited to specific scenarios and data & Real-time feedback integration, real-time learning from human drivers or other professionals \cite{wu2023toward}\cite{gweon2023socially}, crowd-sourced updates \cite{wong2024aligning}, and human-in-the-loop learning \\
\hline

\end{tabular}

\end{table}

\begin{table}[htp]
\centering
\caption{Capabilities and Functionality}
\fontsize{6.5pt}{7.5pt}\selectfont
\label{tab:standards}
\begin{tabular}{|p{1.2cm}|p{1.8cm}|p{2.3cm}|p{2.3cm}|}\hline

\textbf{Feature} & \textbf{Ford SYNC (2007)} & \textbf{MBUX (LLM-based)} & \textbf{Advanced In-Vehicle Voice Assistants (In Research)} \\
\hline
\textbf{Voice Commands} & Specific, pre-programmed commands required & Natural language input, flexible command interpretation & Superior accuracy with emotion detection \cite{zaidi2023cross}, multi-language support, and speaker identification \cite{mamun2024smart} \cite{guan2024integrated}\\
\hline
\textbf{Context Awareness} & None & High contextual awareness, can follow conversation threads & Advanced context awareness, including multi-turn conversations \cite{ma2024multi}\cite{wang2024multi} and understanding of situational context \cite{yang2024situation} \\
\hline
\textbf{Multitasking} & Limited to simple, single tasks & Can handle complex requests and multitasking seamlessly. Multimodal interaction of Voice and touch-based commands & Advanced multitasking \cite{schmidt2023adaptive} with the ability to manage and prioritize multiple unrelated tasks simultaneously.\\
\hline
\textbf{Multimodal Interactions} & None & Multimodal interaction of Voice and touch-based commands & Multimodal Interactions \cite{su2023recent} such as Integration of voice, gestures, facial expressions, and text inputs \\
\hline
\textbf{Personalization} & Minimal to none & Extensive personalization based on user behavior and preferences & Deep personalization \cite{wang2024enhancing} using AI-driven \cite{mehmood2024transformative} user profiling, mood detection, and proactive suggestions \\
\hline

\end{tabular}

\end{table}

\begin{table}[htp]
\centering
\caption{User Interface and Interaction}
\fontsize{6.5pt}{7.5pt}\selectfont
\label{tab:standards}
\begin{tabular}{|p{1.2cm}|p{1.8cm}|p{2.3cm}|p{2.3cm}|}\hline
\textbf{Feature} & \textbf{Ford SYNC (2007)} & \textbf{MBUX (LLM-based)} & \textbf{Advanced In-Vehicle Voice Assistants (In Research)} \\
\hline
\textbf{Interaction Style} & Structured commands only & Conversational, intuitive interactions & Conversational and adaptive, capable of understanding slang, idioms, and varied accents \\
\hline
\textbf{Feedback} & Basic audio feedback, limited adaptability & Dynamic feedback, adapts to user’s preferences and style. Visual and auditory feedback through the infotainment system & Multi-modal feedback with voice, visuals, haptics, and personalized suggestions \\
\hline
\textbf{Ease of Use} & Requires learning specific commands, Requires user learning and memorization of commands & User-friendly, minimal learning curve due to natural language use. Intuitive, with some learning curve & Extremely user-friendly, learns user preferences, and adapts over time \\
\hline

\end{tabular}

\end{table}

\begin{table}[htp]
\centering
\caption{Performance and Efficiency}
\fontsize{6.5pt}{7.5pt}\selectfont
\label{tab:standards}
\begin{tabular}{|p{1.0cm}|p{1.8cm}|p{2.3cm}|p{2.3cm}|}\hline
\textbf{Feature} & \textbf{Ford SYNC (2007)} & \textbf{MBUX (LLM-based)} & \textbf{Advanced In-Vehicle Voice Assistants (In Research)} \\
\hline
\textbf{Accuracy} & Moderate, influenced by user’s command precision. Basic command recognition with potential for errors & High, robust to various accents, dialects, and noisy environments. High accuracy with contextual understanding & Extremely high accuracy with advanced AI, handling complex and ambiguous queries \\
\hline
\textbf{Speed} & Generally fast but can struggle with complex commands due to limited processing power & Extremely fast, handles complex queries efficiently, dependent on cloud connectivity. & Ultra-fast with edge processing and predictive preloading \\
\hline
\textbf{Updates} & Infrequent, requires manual updates. Firmware updates via USB & Continuous updates over the cloud system improves over time. Over-the-air (OTA) updates for software and features & Continuous and automatic OTA updates with AI-driven optimizations \\
\hline

\end{tabular}

\end{table}

\begin{table}[htp]
\centering
\caption{Integration and Extensibility}
\fontsize{6.5pt}{7.5pt}\selectfont
\label{tab:standards}
\begin{tabular}{|p{1.0cm}|p{1.8cm}|p{2.3cm}|p{2.3cm}|}\hline
\textbf{Feature} & \textbf{Ford SYNC (2007)} & \textbf{MBUX (LLM-based)} & \textbf{Advanced In-Vehicle Voice Assistants (In Research)} \\
\hline
\textbf{Service Integration} & Limited to basic phone and media functions & Extensive integration including navigation, internet services, and third-party apps & Seamless integration with a wide range of third-party services and IoT devices \\
\hline
\textbf{Scalability} & Limited scalability, constrained to initial design & Highly scalable, constantly evolving capabilities & Highly scalable across devices and platforms with cloud-edge integration \\
\hline
\textbf{Connectivity} & Limited to in-car systems & Connected to cloud services and smart home devices & Seamless integration across multiple devices, IoT ecosystems, and smart cities \\
\hline
\textbf{Security \& Privacy} & Basic encryption & Advanced encryption with periodic updates & AI-driven threat detection, differential privacy, and adaptive security protocols \\
\hline

\end{tabular}

\end{table}

\subsection{\textbf{Scenario}s}

\begin{table}[h!]
\centering
\caption{Scenario 1: Making a Phone Call, \textbf{Voice Command}: “Call John Smith.”}
\fontsize{6.5pt}{7.5pt}\selectfont
\label{tab:standards}
\begin{tabular}{|p{1.0cm}|p{1.8cm}|p{2.3cm}|p{2.3cm}|}\hline
\textbf{System} & \textbf{Ford SYNC (2007)} & \textbf{MBUX (LLM-based)} & \textbf{Advanced In-Vehicle Voice Assistants (In Research)} \\
\hline
\textbf{Input Requirement} & Exact phrase must be used. & Flexible phrasing like “I need to talk to John Smith.” & \textbf{Highly Flexible Phrasing:} The assistant can understand variations like “Get me John on the line,” “Dial up John Smith,” or even more ambiguous requests like “Can you connect me to John?” The system intelligently parses the intent without requiring a specific phrase. \\
\hline
\textbf{System Output} & “Calling John Smith.” (if command is exact) & “Do you want to call John Smith on mobile or at work?” & \textbf{Personalized Response:}

\begin{itemize}
    \item The assistant might say, “I see you usually call John Smith on his mobile. Should I do that again?”
    \item If the assistant knows your preferences or recent interactions, it may automatically choose the most likely number without asking, but still confirm: “Calling John Smith on his mobile.”
    \item If multiple contacts or numbers are available, the assistant could say, “You have a couple of numbers for John Smith. Would you like to call his mobile, office, or home?”
\end{itemize}
 \\
\hline
\textbf{Context Handling} & None & Understands context; can follow up if multiple numbers exist. & \textbf{Deep Contextual Understanding:}

\begin{itemize}
    \item The assistant remembers your past interactions, such as if you recently spoke to John Smith at a specific time of day or on a specific number. It uses this context to make smart suggestions.
    \item If you’ve mentioned something in an earlier conversation, like “I’ll call John this afternoon,” the assistant might prioritize that number or time.
    \item If you start a follow-up command, like “Actually, text him instead,” the assistant smoothly shifts tasks without losing the context, asking, “Do you want to text John Smith on the same number?”
\end{itemize}
 \\
\hline

\end{tabular}

\end{table}

\begin{table}[h!]
\centering
\caption{Scenario 2: Playing Music, \textbf{Voice Command}: “Play some rock music.”}
\fontsize{6.5pt}{7.5pt}\selectfont
\label{tab:standards}
\begin{tabular}{|p{1.0cm}|p{1.8cm}|p{2.3cm}|p{2.3cm}|}\hline
\textbf{System} & \textbf{Ford SYNC (2007)} & \textbf{MBUX (LLM-based)} & \textbf{Advanced In-Vehicle Voice Assistants (In Research)} \\
\hline
\textbf{Input Requirement} & Must use specific command like “Play genre rock.” & Flexible, understands various phrasings like “Put on some rock.” & The assistant can understand a wide range of natural language requests like “Put on some classic rock,” “Play something by Led Zeppelin,” or even more vague commands like “I want to hear some rock.” The system intelligently infers the user’s intent from context, even if the command isn’t precise. \\
\hline
\textbf{System Output} & If command matches, “Playing rock music.” & “Playing rock from your playlist, or would you like a radio station?” & The assistant might say, “Sure, playing your favorite rock playlist from Spotify.”

If it knows your preferences, it might suggest, “Would you like me to play your ‘Classic Rock Hits’ playlist, or should I create a new station based on rock?”

It could also offer personalized options like, “I’ve found some new rock releases you might like. Should I play those?” \\
\hline
\textbf{Context Handling} & None & Offers choices based on past behavior or available options. & The assistant remembers your previous interactions and adapts to your preferences. For instance, if you often listen to rock in the evening, it might automatically play the songs or artists you typically enjoy at that time.

If you have recently been listening to a specific rock artist or album, the assistant could prioritize that, saying, “Continuing with Led Zeppelin from where you left off?”

The assistant also considers the current environment or situation. For example, if you’re in a quiet environment, it might lower the volume or suggest a softer rock playlist.

If you follow up with a command like, “Actually, I’m in the mood for something heavier,” the assistant smoothly transitions to a different subgenre of rock, like metal, based on your input and past preferences. \\
\hline

\end{tabular}

\end{table}

\begin{table}[h!]
\centering
\caption{Scenario 3: Asking for Weather,\textbf{Voice Command}: “What’s the weather like today?”}
\fontsize{6.5pt}{7.5pt}\selectfont
\label{tab:standards}
\begin{tabular}{|p{1.0cm}|p{1.8cm}|p{2.3cm}|p{2.3cm}|}\hline
\textbf{System} & \textbf{Ford SYNC (2007)} & \textbf{MBUX (LLM-based)} & \textbf{Advanced In-Vehicle Voice Assistants (In Research)} \\
\hline
\textbf{Input Requirement} & May not support weather updates directly. & Understands natural language inquiries. & The assistant can comprehend a wide range of natural language variations, such as “How’s the weather looking?” or “Will I need an umbrella today?” \\
\hline
\textbf{System Output} & “Command not recognized.” & “It’s sunny and 75 degrees today. Do you want the forecast for the week?” & The assistant might say, “It’s currently 75 degrees and sunny. The high today will be 80 degrees, with a slight breeze in the afternoon.”

If it knows your daily routine, it could add, “It should stay sunny for your walk at 5 PM.”

The assistant might also suggest, “Would you like me to set a reminder to bring a jacket this evening when it cools down?” \\
\hline
\textbf{Context Handling} & Cannot handle this request. & Provides detailed response, can extend to related information. & The assistant can automatically adjust its response based on your location and preferences. For instance, if you have an upcoming trip, it might ask, “Would you like to know the weather in New York for your trip tomorrow?”

It could integrate with your calendar and say, “It might rain during your meeting at 3 PM. Would you like me to arrange a ride for you?”

If you have smart home devices, the assistant might suggest, “Should I adjust the thermostat to keep the house cool while it’s warm outside?”

The assistant can also extend the conversation by offering a weekly forecast or details on weather-related alerts, such as “There’s a heat advisory tomorrow. Would you like tips on staying cool?” \\
\hline

\end{tabular}

\end{table}

\begin{table}[h!]
\centering
\caption{Scenario 4: Setting a Destination,\textbf{Voice Command}: “Navigate to the nearest coffee shop.” }
\fontsize{6.5pt}{7.5pt}\selectfont
\label{tab:standards}
\begin{tabular}{|p{1.0cm}|p{1.8cm}|p{2.3cm}|p{2.3cm}|}\hline
\textbf{System} & \textbf{Ford SYNC (2007)} & \textbf{MBUX (LLM-based)} & \textbf{Advanced In-Vehicle Voice Assistants (In Research)} \\
\hline
\textbf{Input Requirement} & Specific commands needed, like “Set destination.” & Flexible, understands various phrasings. & The assistant can interpret various phrasings, such as “Take me to the closest café,” “Where’s the nearest coffee spot?” or even indirect requests like “I need some coffee.” \\
\hline
\textbf{System Output} & May require manual input on the device. & “Found several nearby. Would you like directions to the closest one?” & The assistant might respond, “I found three coffee shops nearby. The closest is Starbucks, which is 0.5 miles away. Would you like to go there, or do you prefer the one with outdoor seating?”

It could also consider your preferences or past behavior, saying, “Would you like to go to the same coffee shop you visited last week?” \\
\hline
\textbf{Context Handling} & Limited to preset commands. & Offers choices, understands "nearest," and provides real-time data. & The assistant can offer real-time updates, such as, “There’s light traffic on the way to Starbucks. Estimated time of arrival is 10 minutes.”

If your vehicle is low on fuel, it might add, “You’re low on gas. Should I find a gas station along the route?”

The assistant can integrate with your calendar and suggest, “You have a meeting in 30 minutes. Would you like me to find a coffee shop closer to your meeting location?”

It can also check for other relevant factors, like “The weather is nice today. Should I find a café with outdoor seating?” \\
\hline

\end{tabular}

\end{table}

\begin{table}[h!]
\centering
\caption{Scenario 5: Asking for next Autonomous Driving Maneuver,\textbf{Voice Command}: “How are you going to handle the upcoming scenario with pedestrians on both sides of the road?” }
\fontsize{6.5pt}{7.5pt}\selectfont
\label{tab:standards}
\begin{tabular}{|p{1.0cm}|p{1.8cm}|p{2.3cm}|p{2.3cm}|}\hline
\textbf{System} & \textbf{Ford SYNC (2007)} & \textbf{MBUX (LLM-based)} & \textbf{Advanced In-Vehicle Voice Assistants (In Research)} \\
\hline
\textbf{Input Requirement} & \textbf{Limited or Unsupported: }Ford SYNC from 2007 would not support such advanced inquiries related to autonomous driving. & \textbf{Natural Language Understanding:} The system can interpret the question due to its LLM-based foundation, understanding the context of autonomous driving. & \textbf{Highly Flexible Understanding:} The assistant can fully comprehend the inquiry, recognizing the complexity of the scenario and the relevance to autonomous driving. \\
\hline
\textbf{System Output} & \textbf{No Response or Error:} The system might respond with “Command not recognized,” or simply provide no response at all. & \textbf{Basic Explanation:} MBUX might respond with a simplified explanation, such as, “I will proceed with caution, maintaining a safe distance from the pedestrians.” & \textbf{Detailed, Real-Time Response:} The assistant could provide a comprehensive response such as, “I will slow down and prepare to stop if necessary. I’m maintaining a safe buffer zone from the pedestrians and monitoring their movements to ensure their safety. The system is also aware of the vehicles around us and will adjust accordingly.” \\
\hline
\textbf{Context Handling} & \textbf{None:} The system lacks any capability to understand or process autonomous driving scenarios. & \textbf{Basic Context Awareness:} The system can provide a general response but may not be able to offer detailed, scenario-specific insights. It might struggle with providing in-depth reasoning or predictions for complex driving situations. & \textbf{Advanced Context Awareness and Predictive Analysis:}

\begin{itemize}
    \item The assistant could offer further details, like, “If any pedestrian steps onto the road, I will initiate an emergency stop. I’m also monitoring traffic behind us to ensure a safe and smooth deceleration.”
    \item It could also integrate with other vehicle systems, providing updates on the road conditions, pedestrian density, and any relevant traffic rules or signals.
    \item The assistant might even suggest, “Would you like me to change lanes to avoid this crowded area altogether?”
\end{itemize}
 \\
\hline

\end{tabular}

\end{table}

\begin{table}[h!]
\centering
\caption{Scenario 6: Asking for upcoming traffic signal status and traffic update, \textbf{Voice Command}: “How much time do you think the current trip will take based on the next traffic signal status?”}
\fontsize{6.5pt}{7.5pt}\selectfont
\label{tab:standards}
\begin{tabular}{|p{1.0cm}|p{1.8cm}|p{2.3cm}|p{2.3cm}|}\hline
\textbf{System} & \textbf{Ford SYNC (2007)} & \textbf{MBUX (LLM-based)} & \textbf{Advanced In-Vehicle Voice Assistants (In Research)} \\
\hline
\textbf{Input Requirement} & \textbf{Limited or Unsupported:} Ford SYNC from 2007 would not support real-time traffic signal status inquiries or dynamic trip time calculations. & \textbf{Natural Language Understanding:} MBUX can understand the question due to its LLM-based foundation, which supports natural language queries related to trip estimation. & \textbf{Highly Flexible Understanding:} The assistant fully understands the complexity of the request, including the need to factor in real-time traffic signal status and traffic data. \\
\hline
\textbf{System Output} & \textbf{No Response or Error:} The system might respond with “Command not recognized” or may provide no response at all. & \textbf{General Estimation:} MBUX might respond with a general trip estimation like, “The estimated time to your destination is 20 minutes. However, I do not have specific data on the upcoming traffic signal status.” & \textbf{Detailed, Dynamic Estimation:} The assistant could provide a precise response such as, “Based on the current traffic signal timing and live traffic conditions, I estimate your trip will take 18 minutes. The next traffic signal is expected to turn green in 30 seconds, which will help reduce your wait time.” \\
\hline
\textbf{Context Handling} & \textbf{None:} The system lacks the capability to process real-time traffic data or predict trip duration based on traffic signals. & \textbf{Limited Context Awareness:} The system can provide an estimated trip time based on current traffic data but does not integrate real-time traffic signal information into its calculations. It may base its estimates on average conditions rather than specific signal timings. & \textbf{Advanced Context Awareness and Real-Time Data Integration:}

\begin{itemize}
    \item The assistant integrates data from traffic signals, live traffic conditions, and historical patterns to deliver a highly accurate trip time estimate.
    \item It could also offer additional insights like, “There’s a slowdown ahead due to construction. Would you like me to suggest an alternate route to save 5 minutes?”
    \item The system might continue to update the estimated time as the trip progresses, adjusting for real-time changes in traffic signals and road conditions.
\end{itemize}
 \\
\hline

\end{tabular}

\end{table}

\begin{figure}
    \centering
     \includegraphics[width=\linewidth,keepaspectratio]{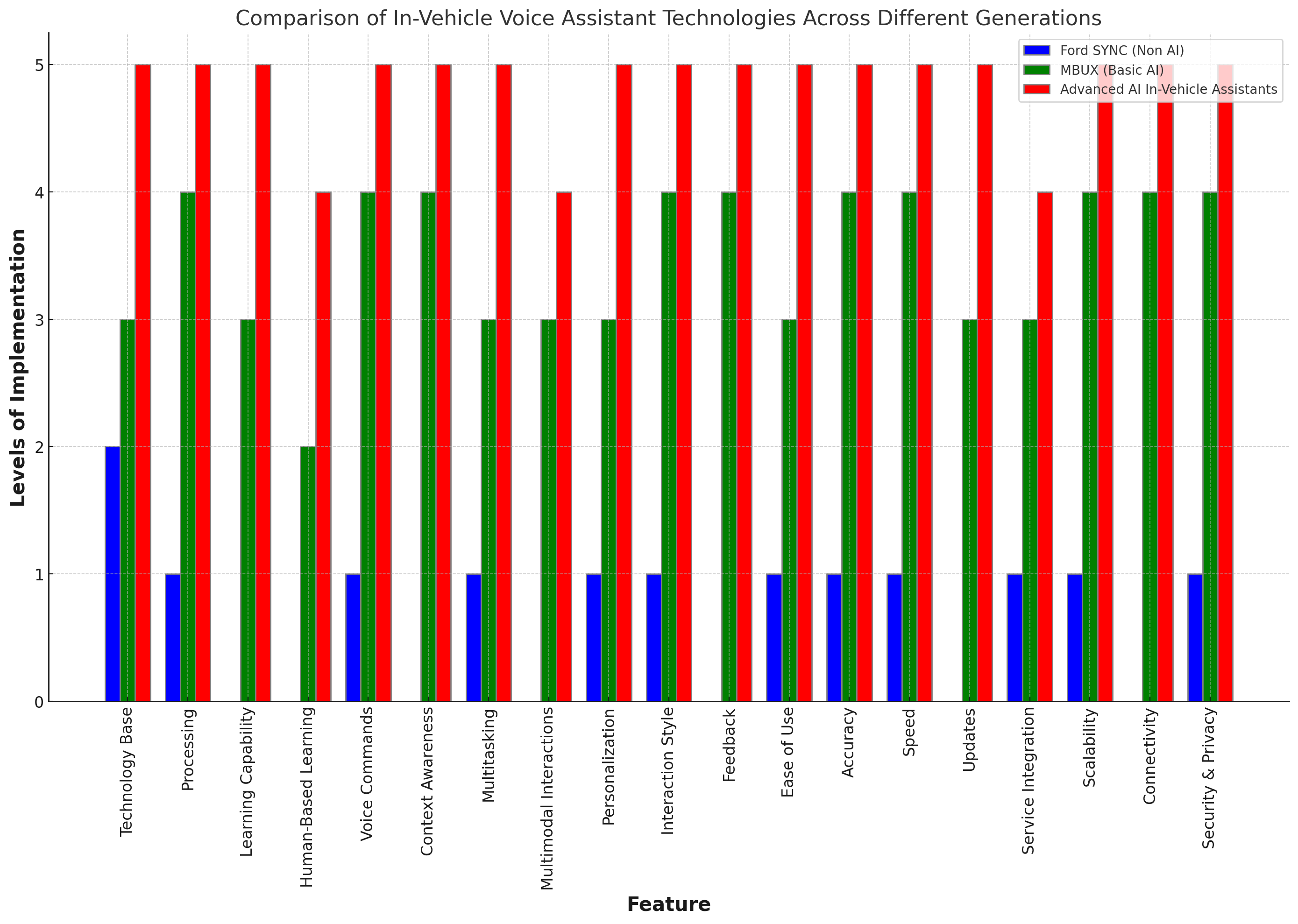}
    \caption{Automotive Voice Assistants Comparison}
    \label{fig:Non-AI and LLM Based Automotive assistants comparison}
\end{figure}

The Fig.\ref{fig:Non-AI and LLM Based Automotive assistants comparison}  compares non-AI automotive assistants (like the 2007 Ford SYNC) with LLM-based automotive assistants (like MBUX) across various features. Each feature is scored out of 5, highlighting the significant advancements that LLM-based systems offer in areas like voice command flexibility, contextual awareness, personalization, integration, and learning capability. As shown, LLM-based systems achieve top scores in all categories, reflecting their more advanced, responsive, and user-centric capabilities.

The Fig.\ref{fig:Non-AI and LLM Based Automotive assistants comparison}  further explores the comparison between non-AI automotive assistants (like Ford SYNC) and LLM-based systems (like MBUX) across different performance-related aspects. Each aspect is rated out of 5, reflecting how each system performs in terms of:

Figures 2 and 3 visualize the summarized performance metrics introduced below.
\begin{itemize}
    \item \textbf{Accuracy}: How accurately the system understands and executes commands.
    \item \textbf{Response Speed}: How quickly the system processes and responds to commands.
    \item \textbf{User-Friendliness}: The ease of interaction and the learning curve for users.
    \item \textbf{Update Frequency}: How often the system receives updates to improve functionality or add features.
    \item \textbf{Noise Handling}: The system's ability to function effectively in noisy environments.
\end{itemize}

LLM-based systems consistently outscore their non-AI counterparts, demonstrating significant improvements in all evaluated aspects, particularly in accuracy, update frequency, and user-friendliness. These advancements underscore the transformative impact of LLM technologies on automotive voice assistants.

\subsubsection{Home voice assistance vs Automotive voice assistance:}

\begin{table}[htp]
\centering
\caption{Home voice assistance vs Automotive voice assistance}
\fontsize{6.5pt}{7.5pt}\selectfont
\label{tab:standards}
\begin{tabular}{|p{1.0cm}|p{1.8cm}|p{2.3cm}|p{2.3cm}|}\hline
\textbf{Feature} & \textbf{Home Voice Assistants} & \textbf{Automotive Voice Assistants} & \textbf{Key Technological Advancements Needed} \\
\hline
 \textbf{Primary Use} & Manage smart home devices, entertainment, and general information & Navigation, vehicle controls, and driver assistance &\textbf{Advanced Integration}: Seamless connectivity with smart home devices or vehicle systems. \\\hline
 \textbf{Operational Conditions} & Stable, indoor environment & Dynamic, often noisy environment with varying connectivity &\textbf{Enhanced Noise Cancellation}: Critical for automotive to handle road and operational noise. \\\hline
 \textbf{Voice Commands} & Broad range including queries, commands, and shopping & Focused on driving-related tasks like navigation and hands-free communication &\textbf{Contextual Command Adjustment}: Adapt voice command processing to context-specific needs. \\\hline
 \textbf{Connectivity} & Dependent on home Wi-Fi and often always connected & Requires cellular connectivity, may have offline capabilities &\textbf{Improved Offline Capabilities}: Essential for automotive systems in areas with poor connectivity. \\\hline
 \textbf{User Interface} & Primarily voice, with some having screens for additional interaction & Voice-driven with integration into vehicle’s display systems &\textbf{Multi-Modal Interaction}: Enhancements in touch and voice integration for both systems. \\\hline
 \textbf{Feedback Mechanism} & Voice, sounds, and visual feedback through LED lights or screens & Voice and visual feedback through the car’s displays &\textbf{Adaptive Feedback Systems}: Dynamic feedback mechanisms that adjust based on the environment and user focus. \\\hline
 \textbf{Profiles} & Multiple user profiles for personalized experiences & Driver profiles that adjust vehicle settings and preferences &\textbf{Cross-Device Personalization}: Advanced profiling that syncs across devices and environments. \\\hline
 \textbf{Learning} & Learns preferences for media, shopping, and routines & Learns preferred routes, climate settings, and common destinations &\textbf{Predictive Modeling Improvements}: Better prediction algorithms for both systems, tailored to specific user behaviors. \\ \hline 
 \textbf{Noise Handling} & Deals with ambient indoor noises & Must handle road noise, music, and multiple speakers simultaneously &\textbf{Sophisticated Acoustic Modeling}: To accurately parse commands in noisy automotive environments. \\ \hline 
 \textbf{Data Sensitivity} & Handles potentially sensitive personal data & Handles sensitive data including location and driving habits &\textbf{Robust Data Encryption}: Ensuring data security, especially in automotive systems handling location and personal info. \\\hline\hline
 \textbf{Security Concerns} & High due to connectivity and access to personal data & High, with added implications for physical safety and vehicle security &\textbf{Advanced Security Protocols}: Increased focus on cybersecurity measures to protect against breaches. \\\hline

\end{tabular}

\end{table}
These six scenarios collectively confirm that generative, multimodal assistants achieve superior contextual reasoning, adaptability, and personalization while maintaining or improving driver safety relative to rule-based predecessors.

\begin{figure}
    \centering
     \includegraphics[width=\linewidth,keepaspectratio]{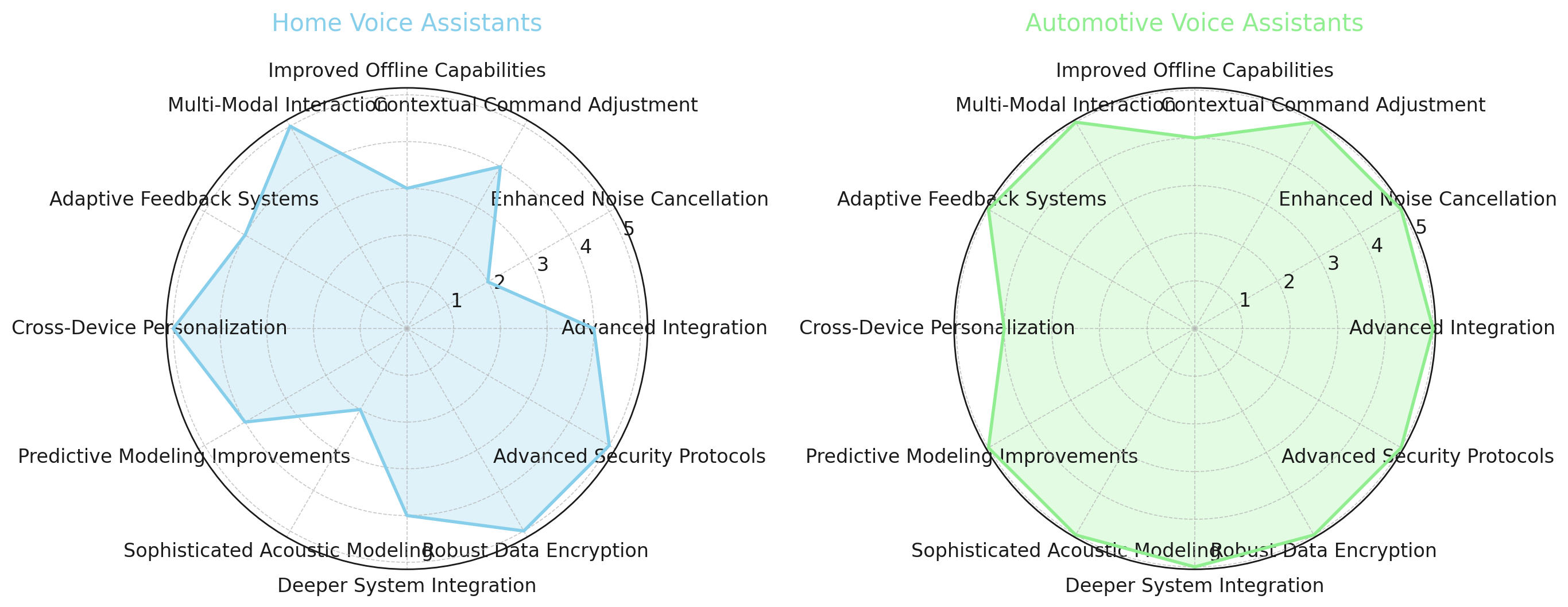}
    \caption{Home vs Automotive Voice Assistants}
    \label{fig:Home Voice Assistant vs Aumotive Voice Assistants}
\end{figure}

The radar in Fig.\ref{fig:Home Voice Assistant vs Aumotive Voice Assistants}  illustrate the key technological advancements needed for LLM-based voice assistant systems in both home and automotive environments. Each chart shows how various technologies are prioritized differently based on their respective settings:

\begin{itemize}
    \item \textbf{Home Voice Assistants} are highlighted in sky blue, showing varying levels of importance across categories, with high emphasis on multi-modal interaction and robust data encryption.
    \item \textbf{Automotive Voice Assistants} are depicted in light green, indicating a generally higher level of importance across all technological aspects, particularly in enhanced noise cancellation, sophisticated acoustic modeling, and advanced security protocols.
\end{itemize}

\section{Future Trends and Research Directions}

Building upon the case study findings, this section outlines anticipated technological and research trajectories for generative AI in the automotive domain. Generative AI is poised to transform the in-vehicle experience, with voice assistance leading the first wave of adoption. Systems such as Mercedes-Benz’s upgraded MBUX Virtual Assistant, built on the MB.OS platform  \cite{heather2024mercedes}, demonstrate how generative models enable more human-like, contextual, and proactive conversations. These assistants are evolving from simple command-and-control interfaces into adaptive copilots that integrate directly with navigation, infotainment, and safety functions. The next stage of progress will emphasize hybrid inference architectures,combining edge computing and cloud processing, to reduce latency while preserving privacy. Advances in automotive-grade SoCs and compact large-language-model frameworks\cite{maginative2024} are making it feasible to deploy multimodal assistants entirely on-device, even in connectivity-constrained environments.

As OEMs integrate these capabilities deeper into their operating systems, assistants will gain persistent context across driving sessions, personalized speech synthesis, and safety-aware task gating. Future versions are expected to act as coordination layers across mobility services booking, parking, payments, and charging, while maintaining compliance with data-protection requirements and driver-distraction regulations \cite{mbusa2024}. In parallel, human-machine-interface research is expanding beyond voice into multimodal interaction that fuses gesture, facial cues, and gaze tracking, enhancing inclusivity and situational awareness.

Beyond in-cabin use, generative AI will increasingly support the wider automotive value chain. In autonomous-driving development, generative world models such as GAIA-1 \cite{hu2023gaia}and diffusion-based simulation engines are accelerating scenario creation for perception and planning validation. By synthesizing rare or hazardous events, these systems expand coverage of the operational-design domain without additional on-road exposure. In design and manufacturing, generative co-pilots are shortening concept-to-prototype timelines by suggesting geometries optimized for aerodynamics, weight, and manufacturability. Major suppliers already employ diffusion and transformer architectures to generate component variations that meet engineering constraints while retaining stylistic consistency \cite{thuraisingham2024trustworthy}. Predictive-maintenance pipelines are also incorporating generative models to simulate degradation patterns, improving early-fault detection and service scheduling across fleets \cite{susmitha2023intricate}

Emerging technologies will reinforce these applications. Diffusion models enhance sensor realism, bridging the sim-to-real gap for camera and radar perception \cite{cakan2024}. Quantized and distilled language models are bringing near-real-time conversational capability to embedded controllers, enabling privacy-preserving operation without cloud dependence \cite{li2024ai}. Generative world models coupled with closed-loop validation frameworks are expected to redefine how safety evidence is collected, allowing thousands of virtual miles to be tested per second with consistent traceability.

Collectively, these advancements signal a fundamental shift in the automotive industry’s digital architecture. Vehicle software stacks are moving from feature-specific modules to integrated ecosystems where generative components create data, scenarios, and user experiences continuously. Cloud-first strategies are giving way to hybrid inference that balances efficiency, security, and regional data sovereignty. Validation workflows are transitioning from static datasets to generative replay and counterfactual analysis, offering richer safety assurance for autonomous and connected vehicles. As generative AI becomes embedded across domains, from cabin interaction to full-vehicle simulation, it will drive a new phase of intelligent mobility characterized by personalization, adaptability, and verifiable safety. Future work should also explore standardization alignment with SAE J3187 (Generative AI System Assurance) and ISO/PAS 8800 (AI safety) to formalize safety evidence for generative components within autonomous-vehicle architectures.

\section{Conclusion}

Generative Artificial Intelligence is emerging as a foundational enabler of the next generation of intelligent, connected, and adaptive vehicles. This review examined how generative models, ranging from GANs and VAEs to diffusion and large-language models, are reshaping design, manufacturing, validation, and in-vehicle user experience. The case study on Mercedes-Benz’s MBUX Virtual Assistant illustrated the rapid transition from rule-based systems to context-aware, conversational copilots that personalize interaction while preserving safety. Across the broader automotive ecosystem, generative world models and synthetic-data frameworks are accelerating autonomous-driving development, while AI-driven co-design and predictive-maintenance tools are shortening product cycles and improving reliability.

At the same time, responsible adoption requires addressing computational cost, data bias, intellectual-property ownership, and the safety assurance of generative outputs. These challenges underline the importance of transparent validation pipelines and standardization efforts such as ISO 21448 (SOTIF) and ISO/PAS 8800 for AI safety. Looking ahead, hybrid edge–cloud deployment, multimodal interfaces, and continual-learning architectures will define the evolution of GenAI in vehicles. If implemented with robust governance and verification, generative AI can transform mobility into an adaptive, self-improving ecosystem that enhances safety, efficiency, and user trust throughout the vehicle lifecycle. With generative AI projected to influence more than 40 percent of in-cabin digital features by 2030 (SAE EDGE Report, 2024), early integration of verifiable AI-safety practices and lifecycle management frameworks will determine the technology’s long-term success and regulatory acceptance.

\textbf{Acknowledgment:}
Please note that this work was performed as part of the IEEE SA P3472 - Standard for Developing Parallel Autonomy Systems \cite{parallela} within Passenger Vehicles which is currently being funded by the IEEE Intelligent Transportation Systems (ITS) society and the Vehicular Technology Society (VTS). As part of this IEEE Standards Association and Committees, there are multiple topics like Intelligent Advanced Voice-Assistants, different control strategies, V2X and other research topics are being studied to enable driving for all. 

\textbf{AI Usage Acknowledgment:}
Please note that AI has been used for better sentence constructs and improving grammar in the paper. AI was not used for any concept, idea or framework generations in this paper. 


%





%


\bibliographystyle{plain}
\bibliography{sample}

%






\end{document}